\documentclass[OTHER]{cta-author}
\usepackage{graphicx,subcaption}
\usepackage{hyperref}
\usepackage{multirow}

\usepackage{xspace}

\begin{document}

\supertitle{Compare transformers and convolutional models}

\title{Comparison between transformers and convolutional models for fine-grained classification of insects}

\author{
	\au{Rita Pucci$^{1}$}
	\au{Vincent J. Kalkman$^{1}$}
	\au{Dan Stowell$^1,^2$}
}

\address{
	\add{1}{Naturalis Biodiversity Center, Leiden(NL)}
	\add{2}{Tilburg University, TSHD, Tilburg(NL)}
	\email{rita.pucci@naturalis.nl}
}

\def\myeffn{\texttt{EffNetv2}\@\xspace}
\def\myvitae{\texttt{ViTAEv2}\@\xspace}
\def\myttv{\texttt{T2TViT14}\@\xspace}
\def\myAE{\texttt{AE}\@\xspace}


\begin{abstract}
Fine-grained classification is challenging due to the difficulty of finding discriminatory features. This problem is exacerbated when applied to identifying species within the same taxonomical class. This is because species are often sharing morphological characteristics that make them difficult to differentiate. We consider the taxonomical class of Insecta. Accurate identification of insects is essential in biodiversity monitoring as they are one of the inhabitants at the base of many ecosystems. Citizen science is doing brilliant work of collecting images of insects in the wild giving the possibility to experts to create improved distribution maps in all countries. Today, we have billions of images that need to be automatically classified and deep neural network algorithms are one of the main techniques explored for fine-grained tasks. At the state of the art, the field of deep learning algorithms is extremely fruitful, so how to identify the algorithm to use? In this paper, we focus on Odonata and Coleoptera orders, and we propose an initial comparative study to analyse the two best-known layer structures for computer vision: transformer and convolutional layers. We compare the performance of T2TViT\_14, a model fully transformer-base, EfficientNet\_v2, a model fully convolutional-base, and ViTAEv2, a hybrid model. We analyse the performance of the three models in identical conditions evaluating the performance per species, per morph together with sex, the inference time, and the overall performance with unbalanced datasets of images from smartphones. Although we observe high performances with all three families of models, our analysis shows that the hybrid model outperforms the fully convolutional-base and fully transformer-base models on accuracy performance and the fully transformer-base model outperforms the others on inference speed and, these prove the transformer to be robust to the shortage of samples and to be faster at inference time.
\end{abstract}

\maketitle



\section{Introduction}
\label{sec:intro}
Fine-grained classification task aims to differentiate between classes that belong to the same superclass. In the biodiversity monitoring field, we talk about species as classes and order as a superclass. The species are usually defined by domain experts (taxonomists) based on morphology or molecular data. The species within an order are closely related and look-alike, i.e. many of them shared colours and characteristics, therefore the fine-grained tasks in this field are particularly challenging. In this paper, we focus on two orders of Insecta: Coleoptera and Odonata in Europe. In the order Coleoptera, there are four main suborders: Archostemata, Myxophaga, Adephaga, and Polyphaga, and more than 130,000 species in Europe~\cite{audisio2015fauna}. The order of Odonata has two main suborders: Epiprocta, and Zygoptera and it is estimated that, in Europe alone, we have more than 200 subspecies. 
\begin{figure}[!b]
\centering{\includegraphics[width=\columnwidth ]{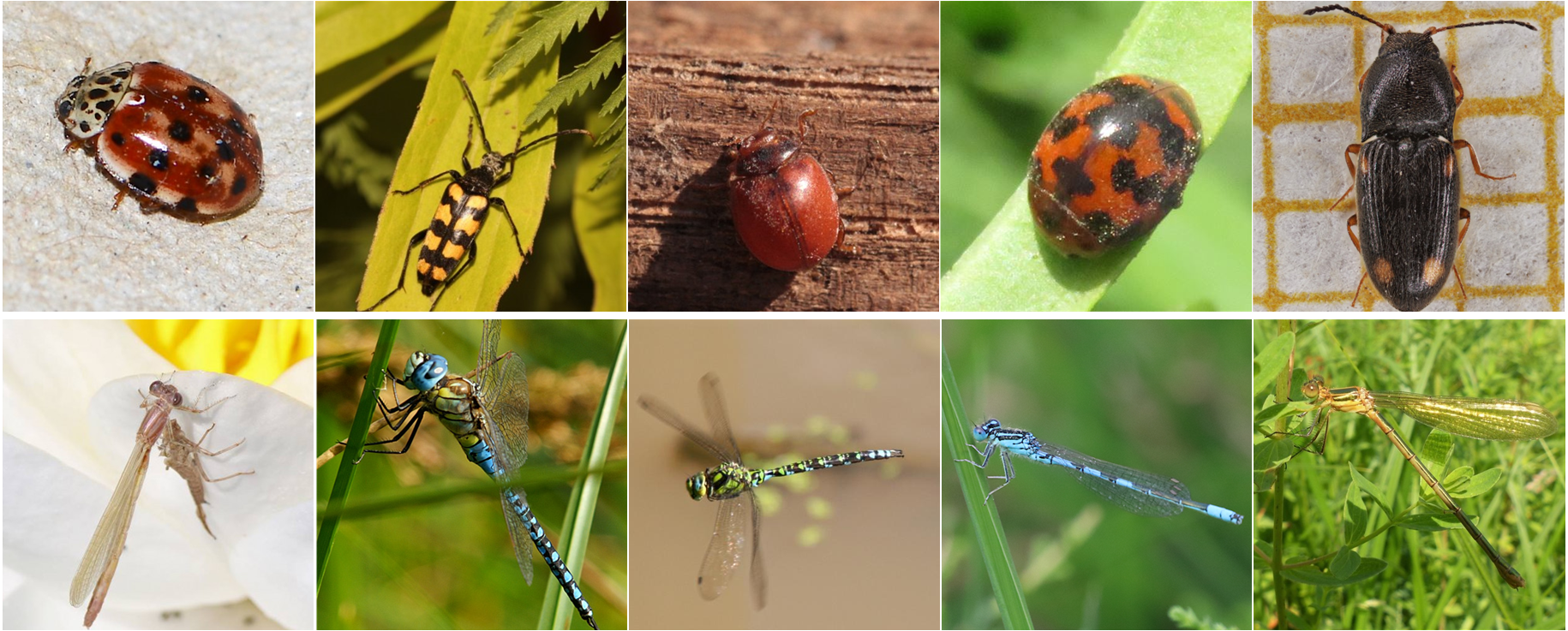}}
\caption{Samples from datasets of Coleoptera and Odonata for fine-grained classification.\label{fig:one}}
\end{figure}
The Coleoptera and Odonata are among the oldest insects with many important roles in our environment but still, they are under-studied. One of the main reasons is the complexity involved in the identification process. In both orders, many species are phenotypically similar in appearance both within and between families. In Coleoptera, many species are small with discriminating characters often difficult to see. A further factor complicating identification is the within-species variability due to differences between life stages, sexes and regional or seasonal variation. In Fig.~\ref{fig:one}, we present some samples from the datasets of interest in this paper. In the first row, all five images are Coleoptera, three of which are from the same family of Coccinellidae but even if they resemble each other to an inexpert eye, they belong to different species. On the second row, we present samples from the order Odonata, as we note the subjects in the images all show a similar shape and dimension and sometimes similar colour, but they all belong to different species. The ability to identify the insects that inhabit the ecosystems is one of the main steps to understanding them.

Despite its significance, the fine-grained task in biodiversity has posed two key challenges: 1) The inter-class variances are often extremely subtle, thus requiring highly discriminative representation for effective classification; 2) As the rarity of the species increases, there are fewer training samples per category, impeding the performance of large-data favoured methods.

The conventional identification technique is to cross-validate the image with the regional field guides, online sources, and field experts. Since these identification methods are highly time-consuming and unaffordable for the common person, there is increasing interest in the investigation of new deep learning fine-grained methods for biodiversity monitoring. Early and fast identification techniques are crucial and the fast-developing of deep learning technologies in computer vision have shown impressive solutions to many real-world problems such as animal identification~\cite{tuia2022perspectives}. 

At the state of the art, the convolutional neural network (CNN) for computer vision is an algorithm based on an inductive bias of locality and shift invariance these two main features make CNN a highly effective deep learning algorithm in image classification. Recently, we see an increased interest in the application of transformers for the same tasks to which CNN was historically devoted. Vision transformer (ViT)~\cite{dosovitskiy2010image} enables multi-head self-attention to capture long-range dependencies within an image and thus can extract diverse feature patterns for discriminative classification. Unfortunately, ViT is data-hungry and  the lack of training data may impede its application in fine-grained tasks. With their pro and cons, both the convolutional and the transformer algorithms are good candidates for fine-grained tasks for insect images but to which extent?

In this paper, we are interested in understanding and evaluating the CNN and the ViT comparing and analysing them in the context of fine-grained tasks in biodiversity monitoring. We consider three of the main families of deep neural networks for computer vision: fully-convolutional, fully-transformer, and hybrid (based on both convolutional and transformer layers). For each of them, we select a model at the state of the art that obtains the best performance in image classification: EfficientNet\_v2~\cite{tan2104efficientnetv2}(\myeffn) for fully-convolutional, T2TViT\_14~\cite{yuan2021tokens}(\myttv) for fully-transformer, and ViTAEv2~\cite{zhang2023vitaev2}(\myvitae) for the hybrid one. For training and validation, we consider datasets collected by citizen science and stored in Observation.org~\cite{Obs}. We evaluate the models on iNaturalist~\cite{iNat} and Artportalen~\cite{art} limited to Odonata and Coleoptera from Europe, which are collected from different communities of citizen science than training. The results are presented to address the fine-grained task at the species and the morph/sex levels. 

\section{Related work}
\label{sec:related_work}
In the domain of Insecta extensive work has been done to identify different species in different orders e.g. Lepidoptera, Coleoptera, Odonata, Orthoptera, and Hymenoptera. In particular, the application of deep learning algorithms such as CNN has seen increased popularity for the ability of automated feature extraction and high accuracy rate in fine-grained classification. 
CNN is now popularly used for insect identification and presents a wide range of models applied to classify Lepidoptera~\cite{theivaprakasham2021identification, chang2017fine, ding2016automatic} which reach high performance in accuracy. Customized models are proposed for generic species from different orders in the class Insecta~\cite{lim2017performance,xia2018insect}, also specifically to classify bees in real time~\cite{dembski2019bees}, Orthoptera for mobile application~\cite{chudzik2020mobile}, and Odonata~\cite{theivaprakasham2022odonata}. 
Even if we have a prolific application of dedicated CNN in fine-grained tasks for insects, these models are not robust in the identification of rare species, resulting in enormous limitations in practical applications. It is still an open challenge that requires investigation. Moreover, we do not observe equal interest in the application of transformer-based models in this task. An interesting comparison between very simple CNN and transformer-based algorithms for fine-grained tasks among species of different kingdoms identifies the ViT model as outperforming the CNN-based models~\cite{peng2022cnn}. A customized transformer model is proposed for insect pest recognition highlighting the need in integrating some of the CNN features into the transformer structure making the model focus more on global coarse-grained information rather than local fine-grained information~\cite{wang2023aa}. 

Though a vast amount of work has been done in the domain of insect identification, to the best of our knowledge and extensive literature survey, we have not found any published research on a comparative evaluation of deep learning models from all three families of deep neural networks for fine-grained identification in biodiversity monitoring. Furthermore, there is no experimentation on the most modern models from computer vision for this task. 

\section{Methods}
\label{sec:length}
\subsection{Models}
\label{sec:models}
We first define the three families of models that we are considering for image classification.
\paragraph{\textbf{Fully-convolutional}} We consider models that are mainly based on the convolutional layers and fully connected layers.
We are interested in models that are competitive with ViT in inference speed, and model size. We choose EfficientNet\_ v2\_ medium (hereafter named as \myeffn)~\cite{tan2104efficientnetv2} for this family of models. The \myeffn has the structure and connections optimised for speed, based on floating point operations per second (FLOPs), and for parameter efficiency and this model represents a good competitor to the transformer-based models. In particular, \myeffn consists of convolutional-based layers~\cite{gupta2019efficientnet, gupta2020accelerator} to better utilise mobile or server accelerators. The CNN models are naturally equipped with intrinsic inductive bias, shift-invariance, and hierarchical structure to extract multi-scale features, and locality. These are proper advantages in extracting representative features from images collected with smartphones. Even if CNN models are commonly used as the backbone of many image classification models at the state of the art, they are not well suited to model long-range dependency due to their structure focused on extracting local features from low level to high level progressively. This can affect the performance in the fine-grained tasks: these models are less inclined to identify relations among details of the subject. The details are typically the characteristics used by taxonomists to distinguish species.

\paragraph{\textbf{Fully-transformer}} Models based only on attention~\cite{vaswani2017attention} are here referred to as fully-transformer models. The Vision Transformer (ViT)~\cite{dosovitskiy2010image} is the first fully-transformer model applied for image classification. ViT demonstrates that transformers are promising for vision tasks. In fact, ViT is based on the self-attention mechanism which allows the model to capture global contextual information, enabling it to learn long-range dependencies and relationships between image tokens (patches). The self-attention mechanism weighs the importance of different tokens in the sequence when processing the input data. In this paper, we consider for comparison a recent evolution of ViT, the Token-2-Token ViT 14~\cite{yuan2021tokens} (\myttv) which uses a progressive tokenization module to aggregate neighbouring tokens into one token. In the first layer, a token is a patch of the image, while in the intermediate layers, a token is a patch of the feature maps. The model is able to extract local information reducing the length of the token iteratively. This architecture reduces the data hunger and boosts the performance relative to the vanilla ViT.

\paragraph{\textbf{Hybrid}} Finally we consider hybrid models, which create a collaboration between the convolutional and the transformer layers. In particular, we consider the \myvitae~\cite{zhang2023vitaev2} which implements inductive bias and the scale-invariance properties into a transformer architecture. To obtain such a result, the algorithm exploits multiple parallel CNN layers for creating the scale invariance and inductive bias, and the transformer layers for creating long-range dependencies among the features extracted.

\section{Experimental configuration}

\subsection{Pre-processing: Augmentation and data preparation}
\label{sec:pre-processing}
For a fair comparison, we implement the same training scheme for all three models. We set the image size as $224\times 224$ and we apply augmentation methods: mixup~\cite{zhang2017mixup}, and cutmix~\cite{yun2019cutmix} for all the models. We do not apply any balancing process and we evaluate the models on the validation split available for the datasets.

\subsection{Dataset}
\label{sec:datasets}
\begin{figure}[!b]
\centering{\includegraphics[width=\columnwidth ]{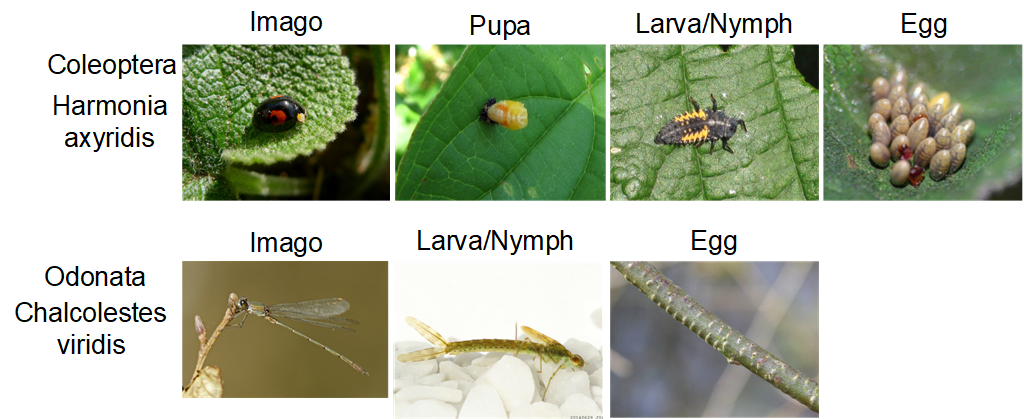}}
\caption{Morphs of a Coleoptera (first row) and an Odonata (second row) species at different life phases.\label{fig:morph}}
\end{figure}

We train all the tree models on the Coleoptera and Odonata datasets from Observation.org~\cite{Obs}, the largest nature platform in the Netherlands for nature observation. Each of these datasets is split into train and test datasets to maximise the number of samples used for training. We validate the models on three datasets: the test split of the Observation.org datasets, the data from iNaturalist.org~\cite{iNat} -the global platform to record and organise nature-related findings-  and from artportalen.org~\cite{art} -the Swedish nature conservation portal. In both orders, we consider only European species. The images are collected with mobile phones by citizen scientists. For the Coleoptera and Odonata datasets from Observation.org, we evaluate the models at the species level and morph level, we also distinguish among samples of morph imago (adults) with different sex (male, female, and unknown). Many species of Coleoptera and all species of Odonata present sexual dimorphism. With species, we refer to the taxonomical full name of the species which consists of the order (e.g. Odonata), the infraorder (e.g. Anisoptera), the family (e.g. Aeshnidae), the genus (e.g. \textit{Aeshna}), and the species (e.g. \textit{affinis}). With morph, we identify specific groups of insects that are all of the same species but differ in morphology e.g. Fig.~\ref{fig:morph} shows a Coleoptera and an Odonata species at different life stages. 
\paragraph{\textbf{Coleoptera dataset}} the dataset, from Observation.org, consists of 849,296 images over 3,087 species. We split the dataset in train and test with the ratio of 80:20 samples per species (674,441:174,855 samples). The dataset is unbalanced with a minimum of 2 samples and a maximum of 11,523 samples per species. The dataset consists of species from 122 families, and we have samples from the Polyphaga and Adephaga suborders and 1,344 genera with a total of 3,087 species. In the dataset, there are samples of thirteen morphs: imago, imago brachypterous, imago macropterous, imago micropterous, unknown, gall, exuviae, deviant, larva/nymph, mine, egg, pupa, queen; and three sexes: male, female, and unknown (Tab.~\ref{tab:samples_ms}). We name this dataset Coleoptera\_obs.

\paragraph{\textbf{Odonata dataset}} the Odonata dataset from Observation.org contains 628,189 images from 235 wild Odonata species. The ratio of train data and test data is roughly 80:20 per species (502,467:125,722 samples). The dataset is unbalanced with a minimum of 2 samples and a maximum of 19,754 samples per species. 
The dataset consists of species from both Epiprocta, in particular from Anisoptera, and Zygoptera infraorders. For the Anisoptera infraorder, we have samples from six families for a total of 153 species, and for the Zygoptera infraorder, we have five families for a total of 82 species, we name this dataset with Odonata\_obs. For this dataset, information is available about morph and sex for further observation in the results. The dataset consists of samples of eight morphs: imago, unknown, fresh imago, exuviae, deviant, larva/nymph, prolarva, egg; and three sexes: male, female, and unknown (Tab.~\ref{tab:samples_ms}).
\begin{table}[!b]
\processtable{Distribution of data at morph/sex level.\label{tab:samples_ms}}
{\begin{tabular*}{20pc}{@{\extracolsep{\fill}}lll@{}}\toprule
Morph and Sex  & Coleoptera\_obs & Odonata\_obs\\
\midrule
imago &130,990&76,025\\ 
imago male&5,963&25,293\\
imago female&5,948&17,756\\
imago unknown&119,079&32,976\\
unknown&31,321&34,675\\
imago macropterous &278&-\\
imago micropterous&9&-\\
imago brachypterous&2&-\\
deviant&2&134\\
fresh imago &-&12,124\\ 
exuviae&16&1,662\\ 
queen&2&-\\
gall&5&-\\
mine&92&-\\
larva/nymph&4,856&1,055\\
pupa&1,019&\\
prolarva&-&4\\ 
egg&87&42\\
\botrule
\end{tabular*}}{}
\end{table}
\paragraph{\textbf{iNaturalist\_Coleoptera}} In the iNaturalist dataset~\cite{iNat_Coleoptera}, there are 236 species of Coleoptera available: among them, 87 species are recognised as European and available in the Coleoptera\_obs dataset train split used in this paper. Our sub-sampled dataset consists of 4,350 samples with a balanced distribution of samples among species (each species has 50 samples). Hereafter, we name this dataset with Coleoptera\_iNat.

\paragraph{\textbf{iNaturalist\_Odonata}} In the iNaturalist dataset~\cite{iNat_Odonata} 291 species are available. Among them, 58 are recognised as European and also available in the Odonata\_obs dataset used for training. The selected 58 species from iNaturalist dataset consist of species from both Anisoptera and Zygoptera infraorders. The dataset has 2,900 samples, 50 samples for each species. Hereafter, we name this dataset with Odonata\_iNat.

\paragraph{\textbf{Artportalen\_Coleoptera}} The Artportalen~\cite{Artportalen_Coleoptera} consists of 3,426 species of Coleoptera. Among them, 1,574 are used to validate the models and are available in the train split of Coleoptera\_obs. The dataset is unbalanced with a total of 118,464 samples. There are more than 400 species with less than 10 samples each and less than 30 species with more than 500 samples each. Hereafter, we name this dataset Coleoptera\_art.

\paragraph{\textbf{Artportalen\_Odonata}} The Artportalen~\cite{Artportalen_Odonata} has 73 species from both Anisoptera and Zygoptera infraorders of which 69 species are available in the train split of Observation.org. The dataset consists of 55,680 samples and it is unbalanced with 12 species with less than 100 samples and 20 species with more than 1,000 samples. Hereafter, we name this dataset with Odonata\_art.

\subsection{Model settings and hyperparameters}
All three selected models are pre-trained on ImageNet1k~\cite{deng2009imagenet} and fine-tuned on the train split of the datasets. We modify the linear layer of the head of the original structures (the classifier layer) to be in line with the number of classes required for our datasets. Models are implemented on Pytorch Image Models (timm)~\cite{rw2019timm} and executed on NVIDIA A40 GPU.
For the models trained on Coleoptera\_obs, we trained the model for a maximum of 90 epochs. Due to the high number of species and the low number of samples per species, we apply early stopping regularisation based on training loss to avoid overfitting, this is because we do not use a validation split. In this case, the \myvitae model stopped at 31 epochs, the \myttv model stopped at 89 epochs, and the \myeffn stopped at 66 epochs. 
For the models trained on Odonata\_obs, we trained the models for 61 epochs and we did not apply early stopping, because we did not observe overfitting behaviour with this dataset. 
With both datasets, we used a batch size of 32 samples, $5 \times 10^{-4}$ as the learning rate, 0.065 weight decay, with  AdamW~\cite{kingma2014adam} as the optimiser with cosine learning rate decay~\cite{loshchilov2016sgdr}, and 10 warm-up epochs.

\section{Experimental results}
\label{sec:res}
The three models are fine-tuned on the Coleoptera\_obs and the Odonata\_obs datasets train split. We evaluate the models on the Coleoptera and the Odonata datasets from Observation.org and from iNaturalist and Artportalen sub-datasets. 
To evaluate the models, we consider top-1 accuracy, hereafter named top-1, the model prediction with the highest probability must be exactly the expected answer; top-5 accuracy, hereafter named top-5, which considers any of our models' top 5 highest probability answers match with the expected answer. Moreover, we evaluate the range distribution of accuracy per species, the accuracy among species, the F1score, and the inference speed. We then examine the performance of the models considering the morph and sex in the case of insects at the morph imago. 

\paragraph{\textbf{Results based on species}}
\label{par:species}
\begin{table}[!b]
\processtable{The top-1, top-5, and F1score for the fine-grained task at species level. In the first block, the models are trained on the train split of the Coleoptera\_obs dataset and tested on the test split of the Coleoptera\_obs dataset and on the Coleoptera\_iNat and Coleoptera\_art datasets. The results are to be considered for the Odonata training in the second block of rows.\label{tab1}}
{\begin{tabular*}{20pc}{@{\extracolsep{\fill}}lllll@{}}\toprule
Dataset  &Model  & top-1 & top-5 & F1score\\
\midrule
 \multirow{3}{*}{Coleoptera\_obs} &\textbf{\myvitae  }&\textbf{89.8\%} & \textbf{97.5\%} & \textbf{89.53\%} \\
&\myeffn &88.0\% & 96.7\% & 87.78\% \\
&\myttv  &88.1\% & 96.7\% & 87.97\% \\
\\
\multirow{3}{*}{Coleoptera\_iNat} &\textbf{\myvitae} & \textbf{83.6\%} &\textbf{92.4\% }&\textbf{81.26\%} \\
&\myeffn  &78.6\% & 90.0\% &78.56\% \\
&\myttv & 78.9\% & 90.6\% & 76.64\% \\
\\
\multirow{3}{*}{Coleoptera\_art} &\textbf{\myvitae} & \textbf{90.4\%} &\textbf{96.7\% }&\textbf{90.44\%} \\
&\myeffn  &88.8\% & 96.1\% &88.55\% \\
&\myttv & 87.3\% & 96.2\% & 87.42\% \\

\midrule
 \multirow{3}{*}{Odonata\_obs} &\textbf{\myvitae}&\textbf{93.6\%}   &\textbf{98.7\%}& 93.29\% \\
&\myeffn  &92.6\% &98.5\% &\textbf{94.30\%} \\
&\myttv &91.5\% &98.4\% & 93.65\% \\
\\
 \multirow{3}{*}{Odonata\_iNat} &\textbf{\myvitae} &81.4\%&\textbf{91.2\%}&\textbf{84.37\%} \\
&\textbf{\myeffn} &\textbf{82.4\%} &\textbf{91.2\%} &80.03\%\\
&\myttv & 79.8\% & 90.2\% & 74.56\% \\
\\
 \multirow{3}{*}{Odonata\_art} &\textbf{\myvitae} &\textbf{69.2\%}&\textbf{84.8\%}&\textbf{68.25\%} \\
&\myeffn &67.3\% &82.9\% &65.99\%\\
&\myttv & 66.7\% & 83.4\% & 65.72\% \\
\botrule
\end{tabular*}}{}
\end{table}

We evaluate the three models computing an overall top-1 and top-5 averaging the results among the species. Tab.~\ref{tab1} shows the results obtained in the evaluation with the validation datasets. The Coleoptera\_obs and Coleoptera\_art datasets are heavily unbalanced with a high number of species and a low number of samples. For these datasets, the performance in all three metrics (top-1, top-5, and F1score) is in favour of \myvitae which shows to be robust on different distributions as proven by the fact that similar behaviour is observable also with the results on Coleoptera\_iNat. The Odonata\_obs and Odonata\_art datasets also are unbalanced but they consist of a high number of samples and a low number of species. The models perform almost equally on both the Odonata\_obs and Odonata\_iNat datasets, while with Odonata\_art the performance in top-1 accuracy is lower than 70\%. Overall, the models have equal performance in top-5 accuracy with all three datasets, we take this as confirmation of the robustness of the models. With Odonata\_obs and Odonata\_iNat. The \myvitae outperforms the others in all three metrics for almost all the datasets. We can conclude that the \myvitae can be a better candidate compared to the other two models on the average species accuracy.
\begin{figure*}[!b] 
\begin{subfigure}{.25\textwidth}
  \centering
  \includegraphics[width=\linewidth]{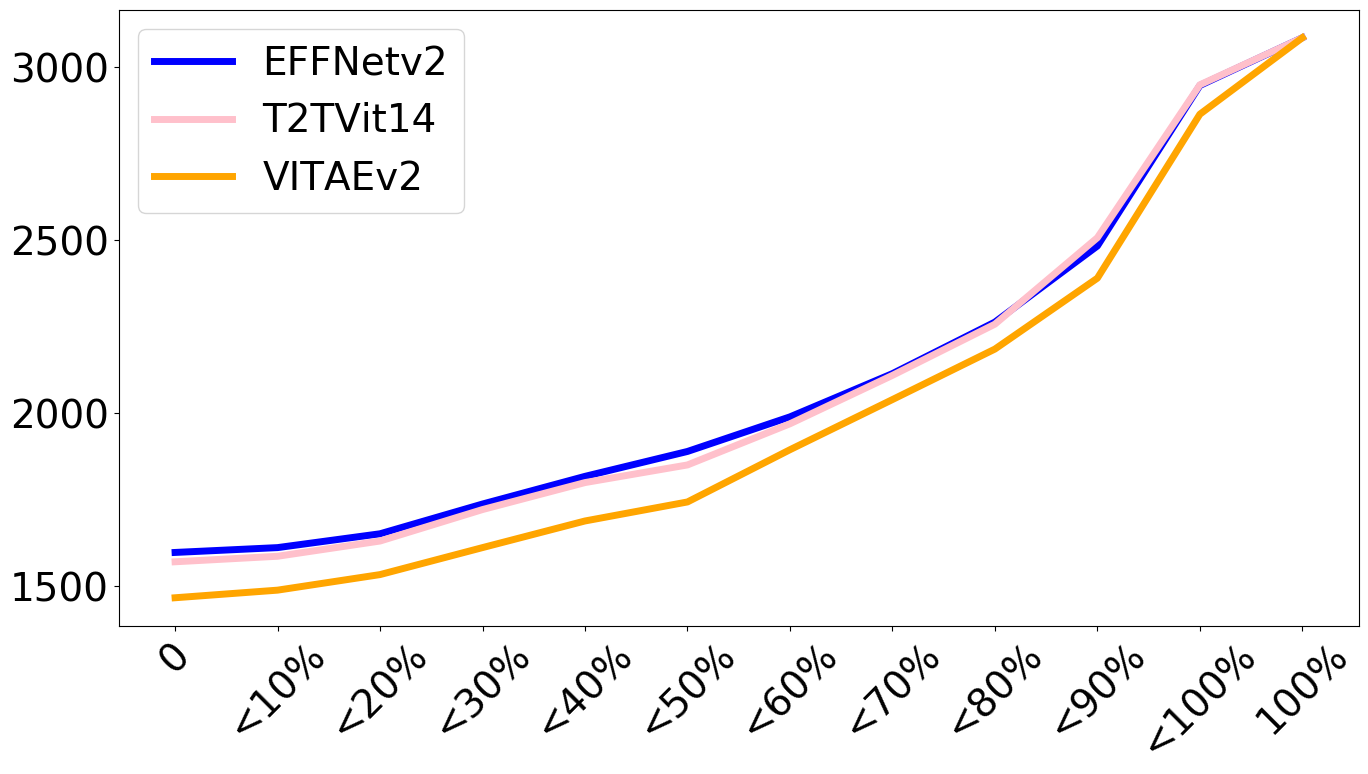}
  \caption{Coleoptera\_obs: top-1}
  \label{fig:clacc1}
\end{subfigure}%
\begin{subfigure}{.25\textwidth}
  \centering
  \includegraphics[width=\linewidth]{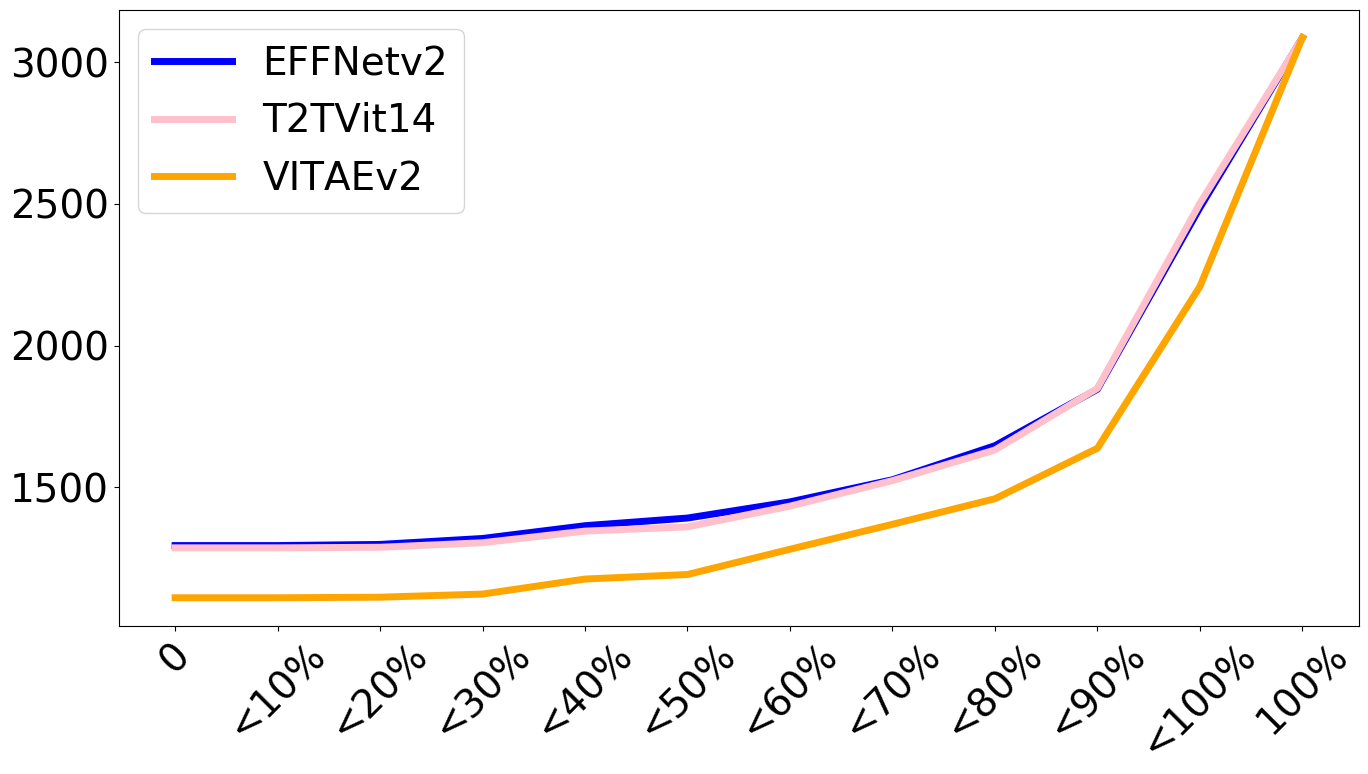}
  \caption{Coleoptera\_obs: top-5}
  \label{fig:clacc5}
\end{subfigure}
\begin{subfigure}{.25\textwidth}
  \centering
  \includegraphics[width=\linewidth]{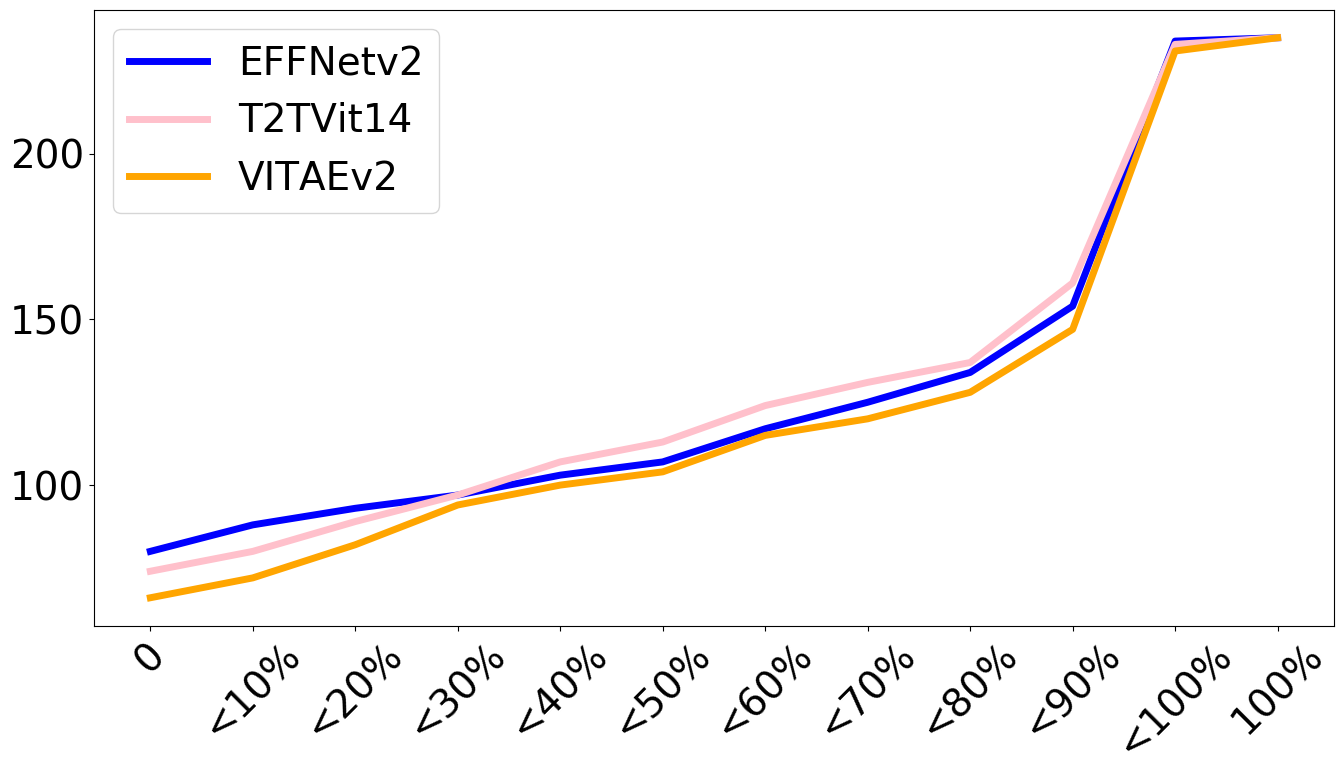}
  \caption{Odonata\_obs: top-1}
  \label{fig:odacc1}
\end{subfigure}
\begin{subfigure}{.25\textwidth}
  \centering
  \includegraphics[width=\linewidth]{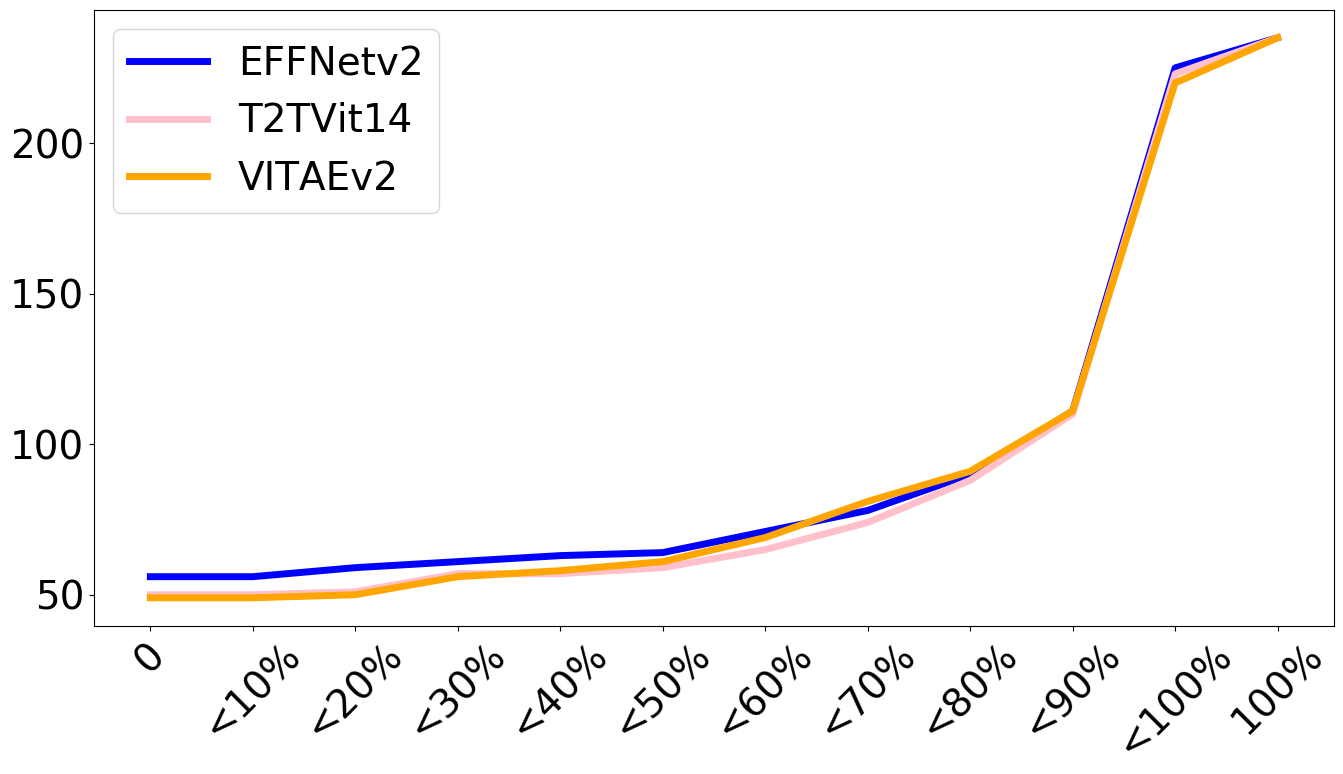}
  \caption{Odonata\_obs: top-5}
  \label{fig:odacc5}
\end{subfigure}
\begin{subfigure}{.25\textwidth}
  \centering
  \includegraphics[width=\linewidth]{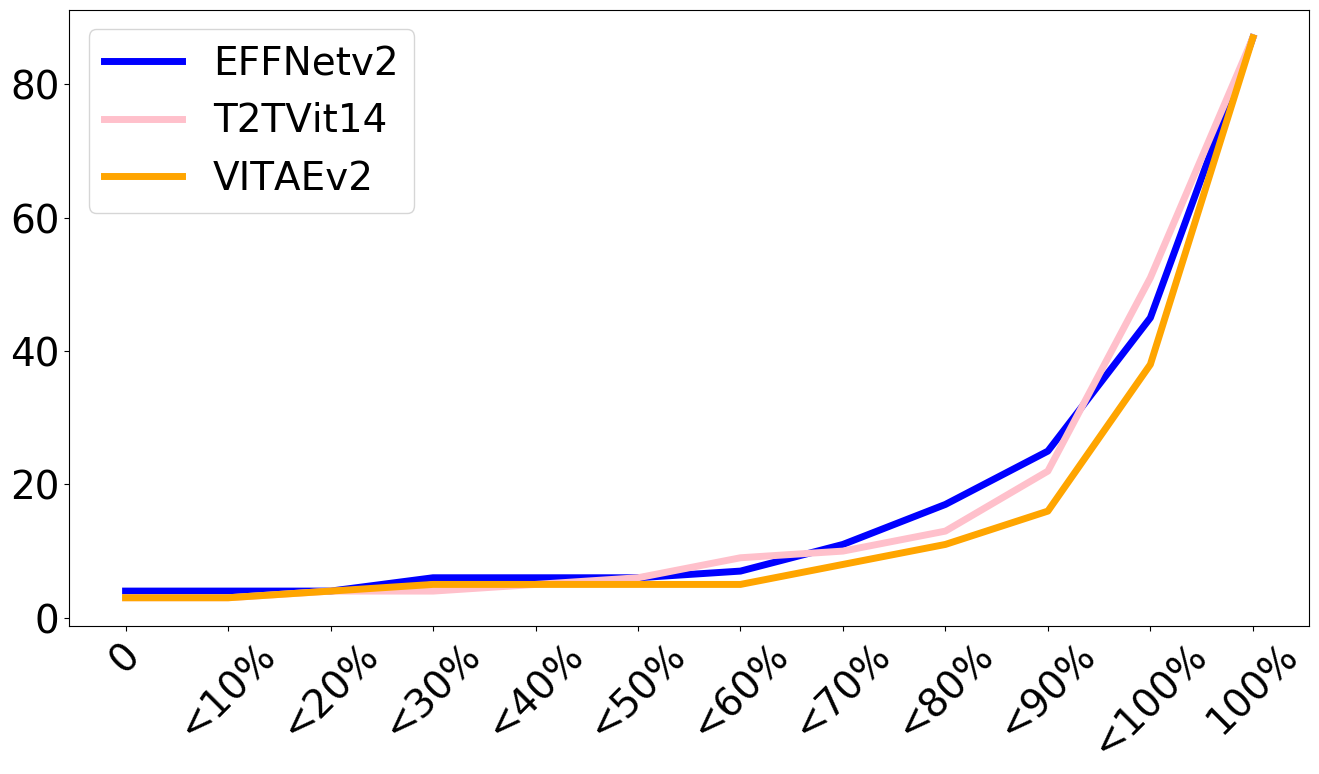}
  \caption{Coleoptera\_iNat: top-1}
  \label{fig:clacc1inat}
\end{subfigure}%
\begin{subfigure}{.25\textwidth}
  \centering
  \includegraphics[width=\linewidth]{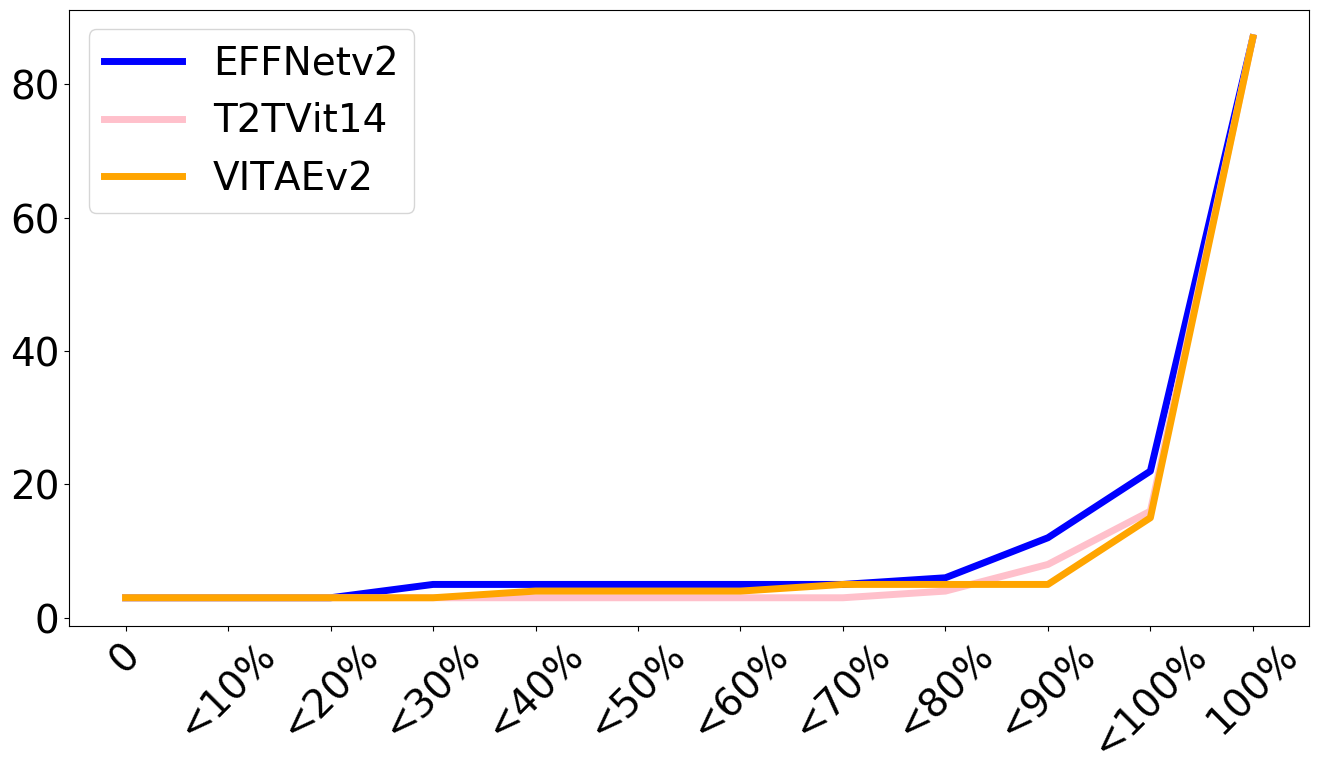}
  \caption{Coleoptera\_iNat: top-5}
  \label{fig:clacc5inat}
\end{subfigure}
\begin{subfigure}{.25\textwidth}
  \centering
  \includegraphics[width=\linewidth]{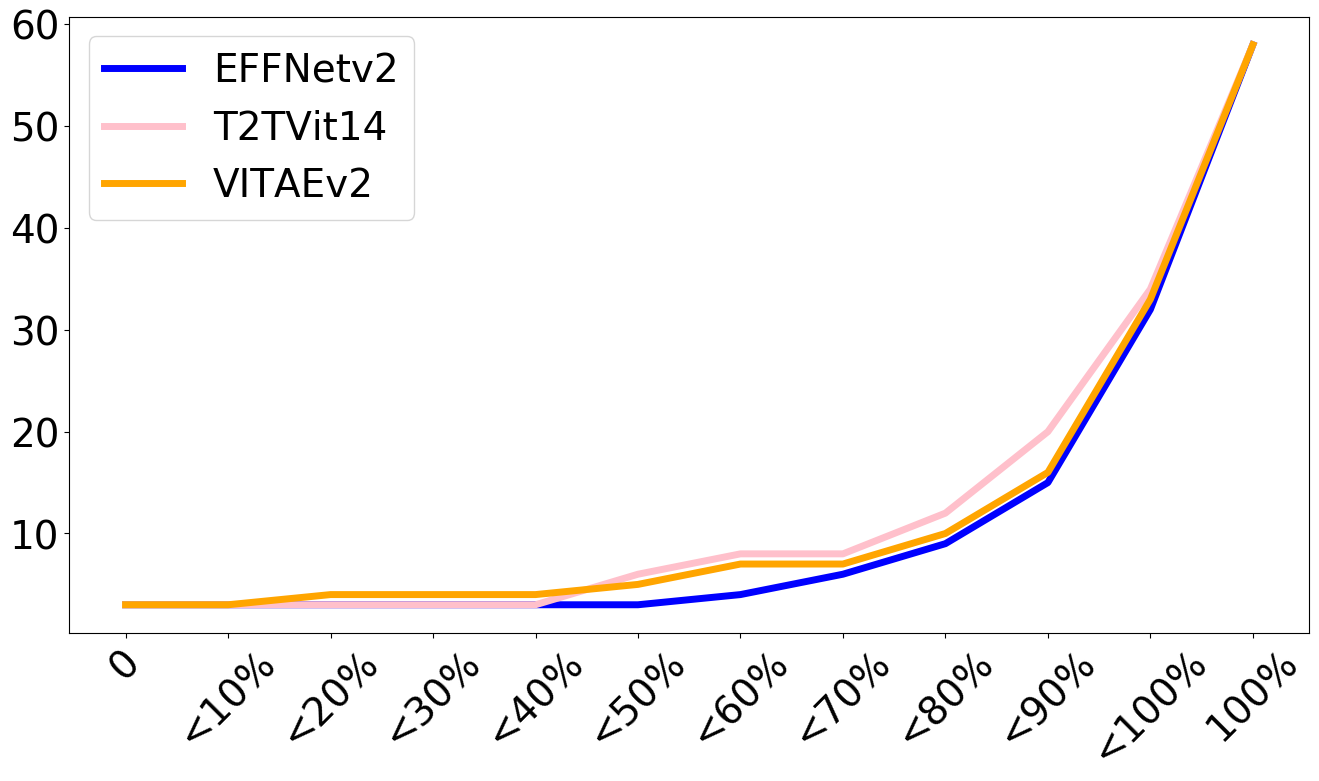}
  \caption{Odonata\_iNat: top-1}
  \label{fig:odacc1inat}
\end{subfigure}
\begin{subfigure}{.25\textwidth}
  \centering
  \includegraphics[width=\linewidth]{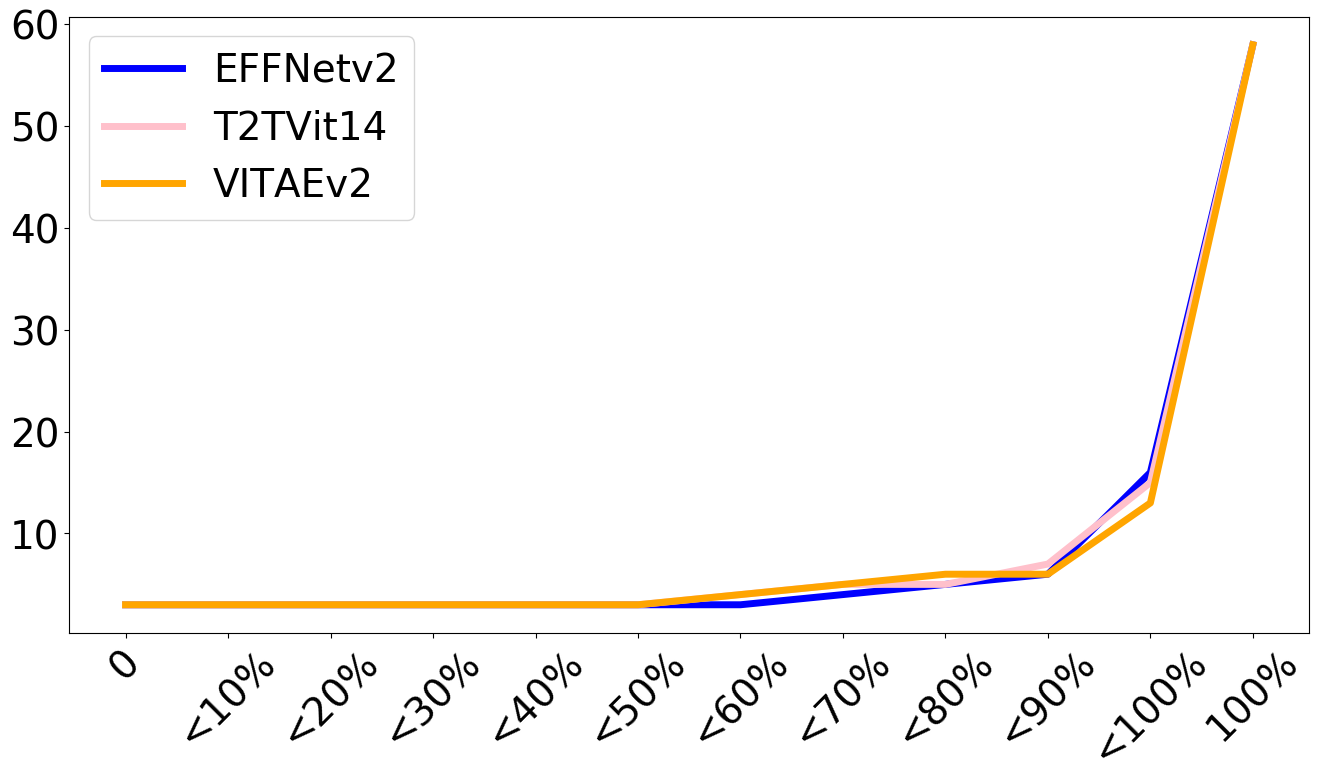}
  \caption{Odonata\_iNat: top-5}
  \label{fig:odacc5inat}
\end{subfigure}
\begin{subfigure}{.25\textwidth}
  \centering
  \includegraphics[width=\linewidth]{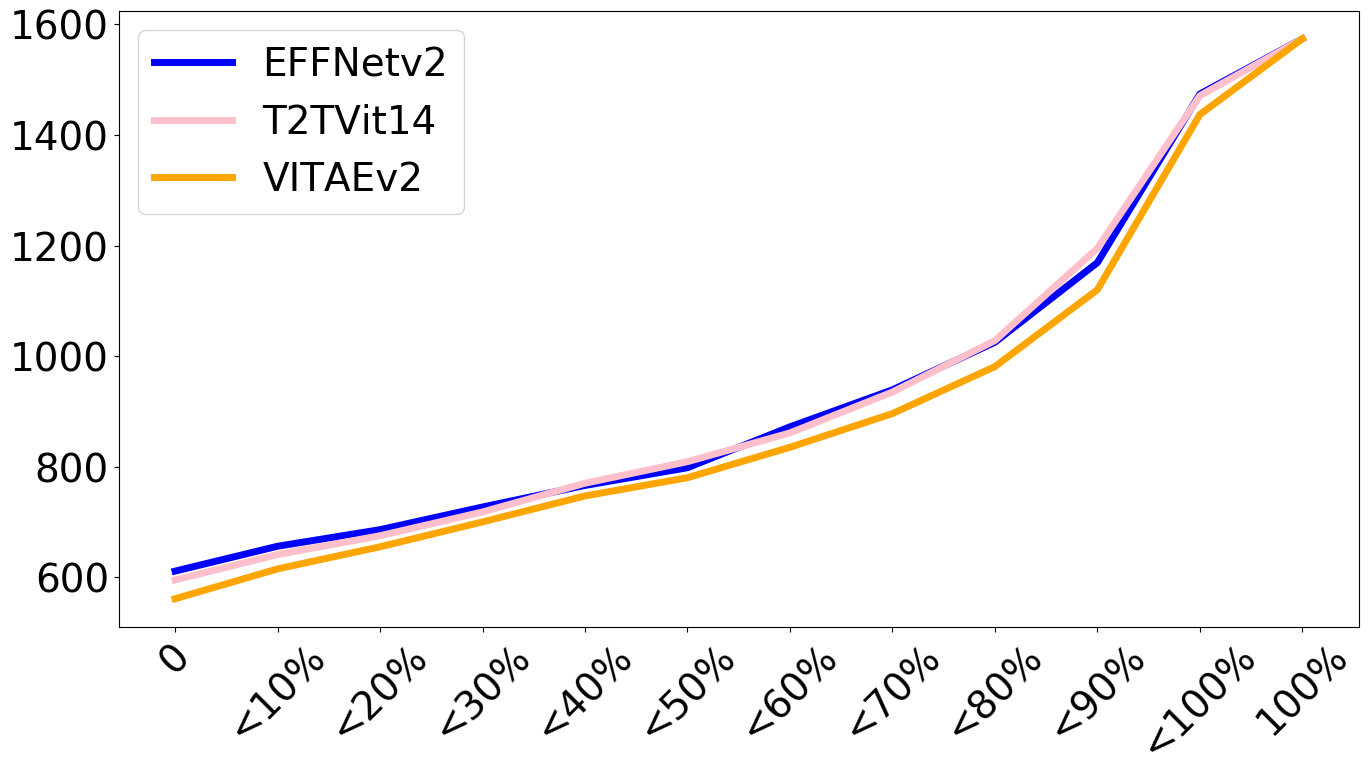}
  \caption{Coleoptera\_art: top-1}
  \label{fig:clacc1art}
\end{subfigure}%
\begin{subfigure}{.25\textwidth}
  \centering
  \includegraphics[width=\linewidth]{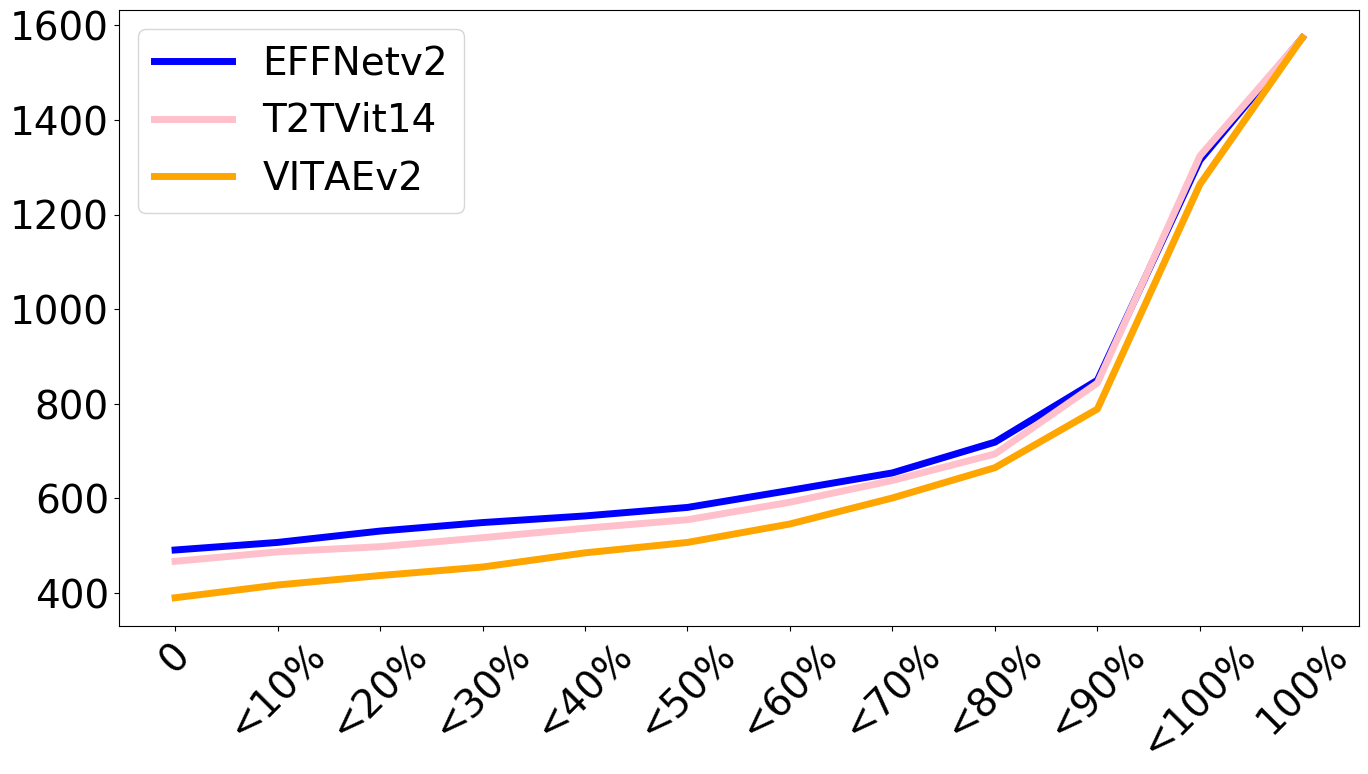}
  \caption{Coleoptera\_art: top-5}
  \label{fig:clacc5art}
\end{subfigure}
\begin{subfigure}{.25\textwidth}
  \centering
  \includegraphics[width=\linewidth]{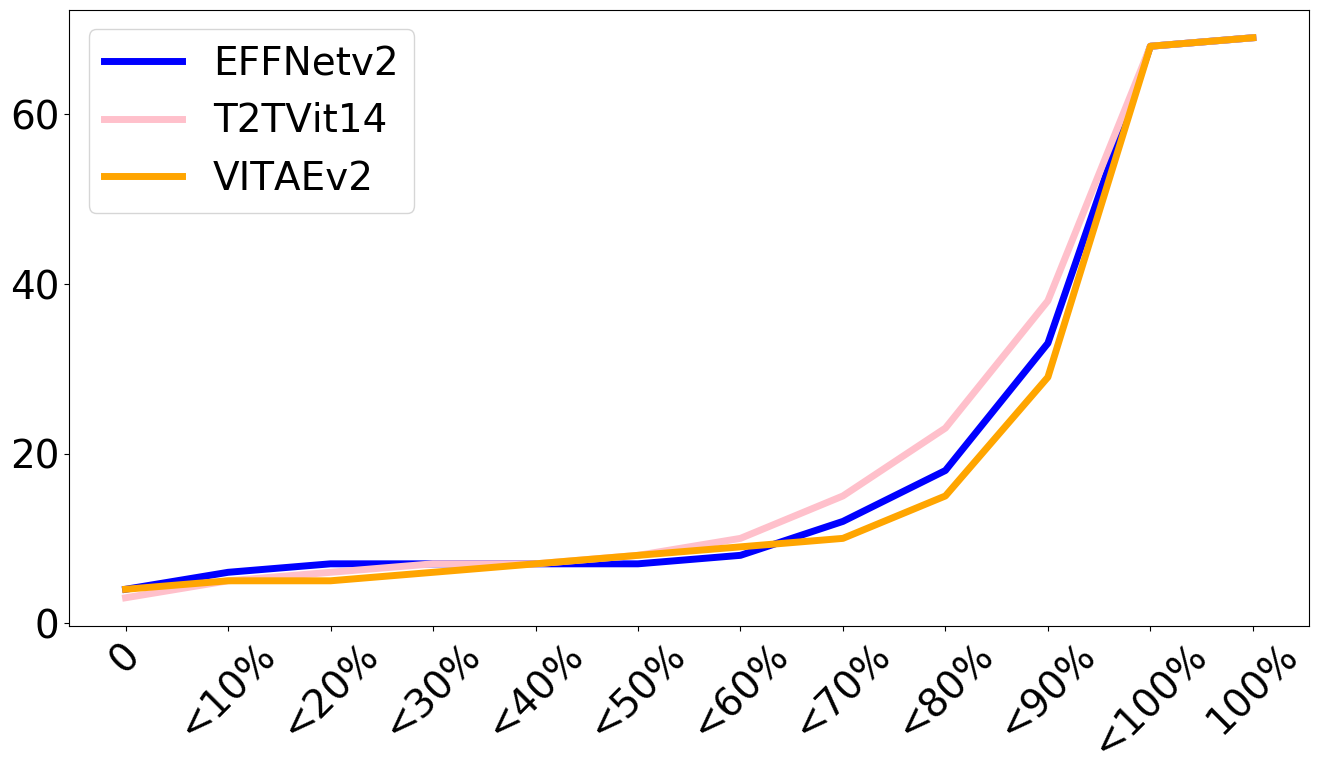}
  \caption{Odonata\_art: top-1}
  \label{fig:odacc1art}
\end{subfigure}
\begin{subfigure}{.25\textwidth}
  \centering
  \includegraphics[width=\linewidth]{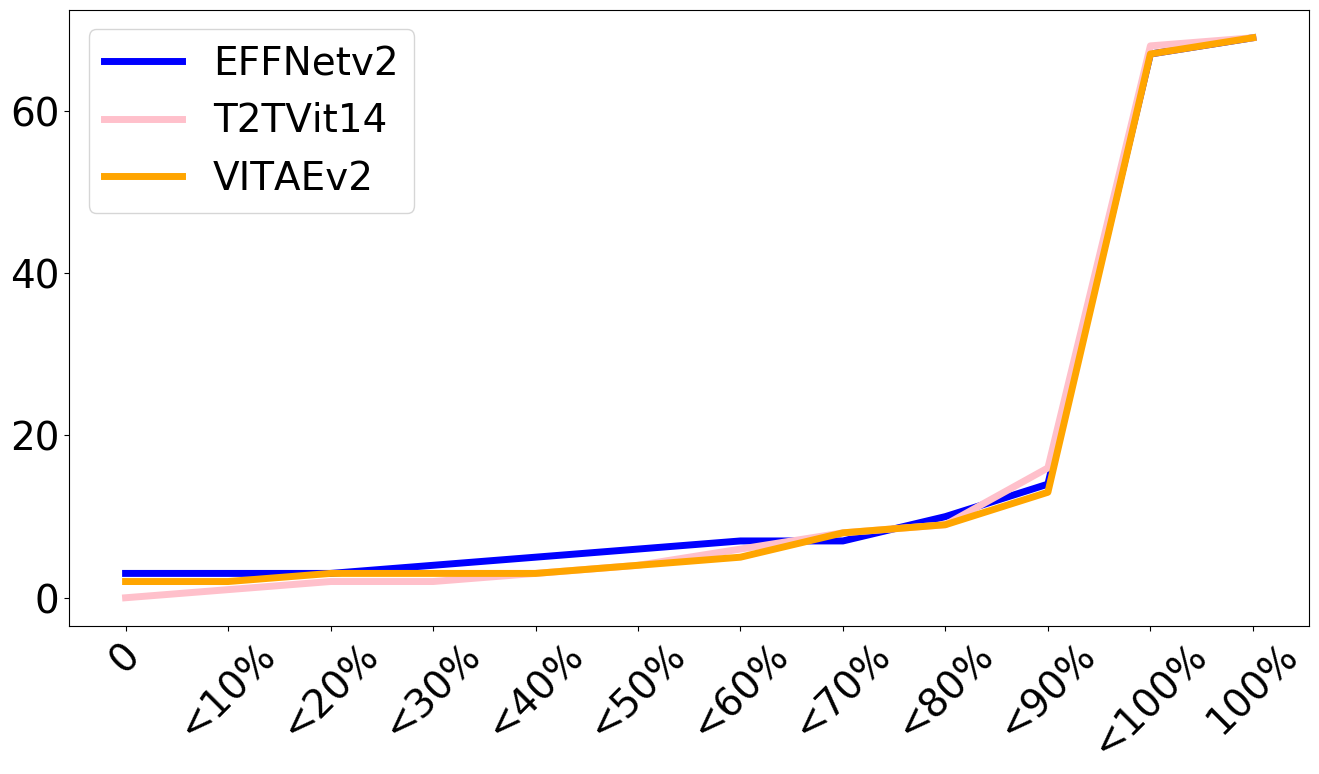}
  \caption{Odonata\_art: top-5}
  \label{fig:odacc5art}
\end{subfigure}
\caption{The cumulative distribution of top-1 and top-5 accuracy per species. The x-axis is the range of accuracy, and on the y-axis is the number of species, i.e. in Fig.(a) the model \myvitae has accuracy $<10\%$ for less than 1500 species and accuracy $>90\%$ for more than 500 species.}
\label{fig:accClOd}
\end{figure*}

In the per-species accuracy (Fig.~\ref{fig:accClOd}), the first line of plots shows the results obtained with the Coleoptera\_obs and the Odonata\_obs datasets, the second and third lines show the results obtained with the iNaturalist and Artportalen respectively. We observe that with both the Coleoptera\_obs and the Odonata\_obs datasets and in both top-1 and top-5 accuracy, \myvitae presents a lower amount of species with $0\%$ accuracy than the other two considered models, and a higher amount of species with accuracy higher than $80\%$. A similar behaviour is shown also for iNaturalist and Artportalen datasets, though with iNaturalist, both \myvitae and \myttv obtain accuracy higher than $90\%$ for more species than \myeffn, while with Artportalen, \myvitae outperforms the other models.

\begin{figure*}[!b]
\centering
\begin{subfigure}{\linewidth}
  \centering
  \includegraphics[width=\linewidth]{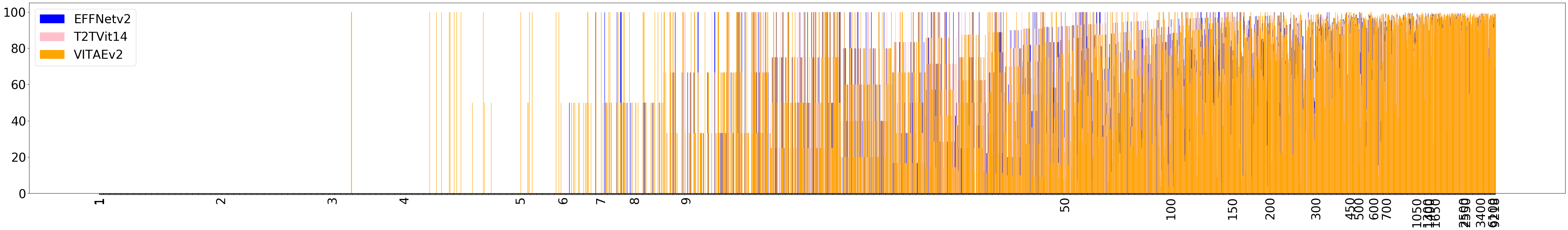}
  \caption{Coleoptera\_obs}
  \label{fig:clacc1}
\end{subfigure}
\begin{subfigure}{\linewidth}
  \centering
  \includegraphics[width=\linewidth]{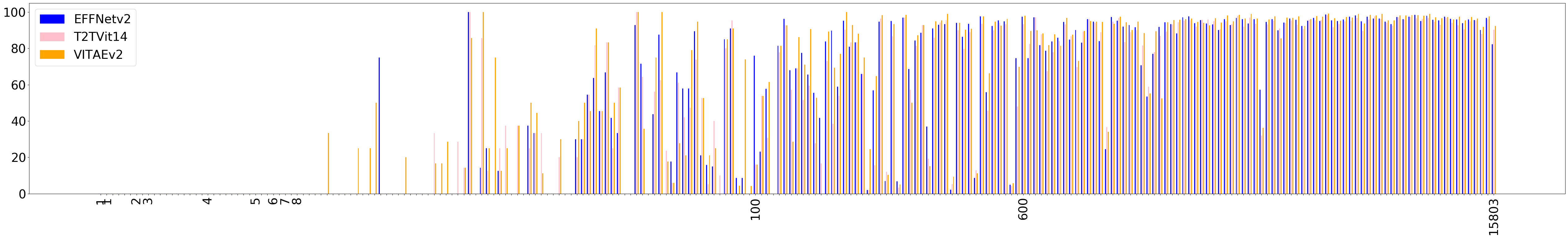}
  \caption{Odonata\_obs}
  \label{fig:odacc1}
\end{subfigure}
\caption{Performance in top-1 accuracy obtained at species level sorted by the number of training samples for each species.}
\label{fig:accSpecies}
\end{figure*}

We now consider the relation between the amount of sample available per species in the train split and the accuracy top-1 obtained by the models with the test split. Fig.~\ref{fig:accSpecies} shows the top-1 accuracy of the models per species, on the x-axis is the number of samples available in the train split for the species, and on the y-axis is the accuracy obtained. We observe that the \myvitae is the only model able to predict species with less than 5 samples in train split within $50\% - 100\%$ top-1, and all three models obtain top-1 higher than $50\%$ for species with more than 7 samples in train split. For species with more than 45 samples in train split, the three models reach between $80\% - 100\%$ top-1. This behaviour is observed for the Coleoptera\_obs and for the Odonata\_obs which require more than 100 samples. We conclude that the ability of \myvitae to learn the singularity of each species based on details of the images is stronger than fully-convolutional and fully-transformer based models. That is shown by its ability to identify species even with few samples in training.
\begin{table}[!b]
\processtable{Table shows the inference speed (sec/sample) of the models trained on Odonata\_obs and on Coleoptera\_obs.\label{tab2}}
{\begin{tabular*}{20pc}{@{\extracolsep{\fill}}lll@{}}\toprule
model  & Coleoptera\_obs & Odonata\_obs \\
\midrule
\myvitae  &0.041 ($\pm 0.002$) & 0.0185 ($\pm 0.0115$) \\
\myeffn  &0.039 ($\pm 0.003$)& 0.0115 ($\pm 0.0065$) \\
\textbf{\myttv} &  \textbf{0.037} ($\pm 0.001$)& \textbf{0.0090} ($\pm 0.0040$) \\
\botrule
\end{tabular*}}{}
\end{table}

\paragraph{Inference speed} We evaluate the two taxa (Coleoptera and Odonata) separately because the difference in the output dimension impacts the number of parameters at the head layer thus at the computing time. The models' inference speed is presented in Tab.~\ref{tab2}. On the Coleoptera\_obs, all three models achieve a similar average inference speed per sample around $0.039sec$. On the Odonata\_obs, we observe that the \myttv outperforms the other two models providing the prediction per sample on an average time of $0.009sec$. 
To give the reader the impact of the inference speed, to infer the prediction of the entire test split: with the Coleoptera\_obs dataset, \myttv takes $2.19hrs$, \myeffn takes $2.29hrs$, and \myvitae takes $2.39hrs$; with the Odonata\_obs dataset, \myttv takes $18min51sec$, \myeffn takes  $24min5sec$, and \myvitae takes $38min45sec$.

\paragraph{\textbf{Results based on Morph and sex}}
\label{par:morphsex}
\begin{figure*}[!b]
\begin{subfigure}{.5\textwidth}
  \centering
  \includegraphics[width=\linewidth]{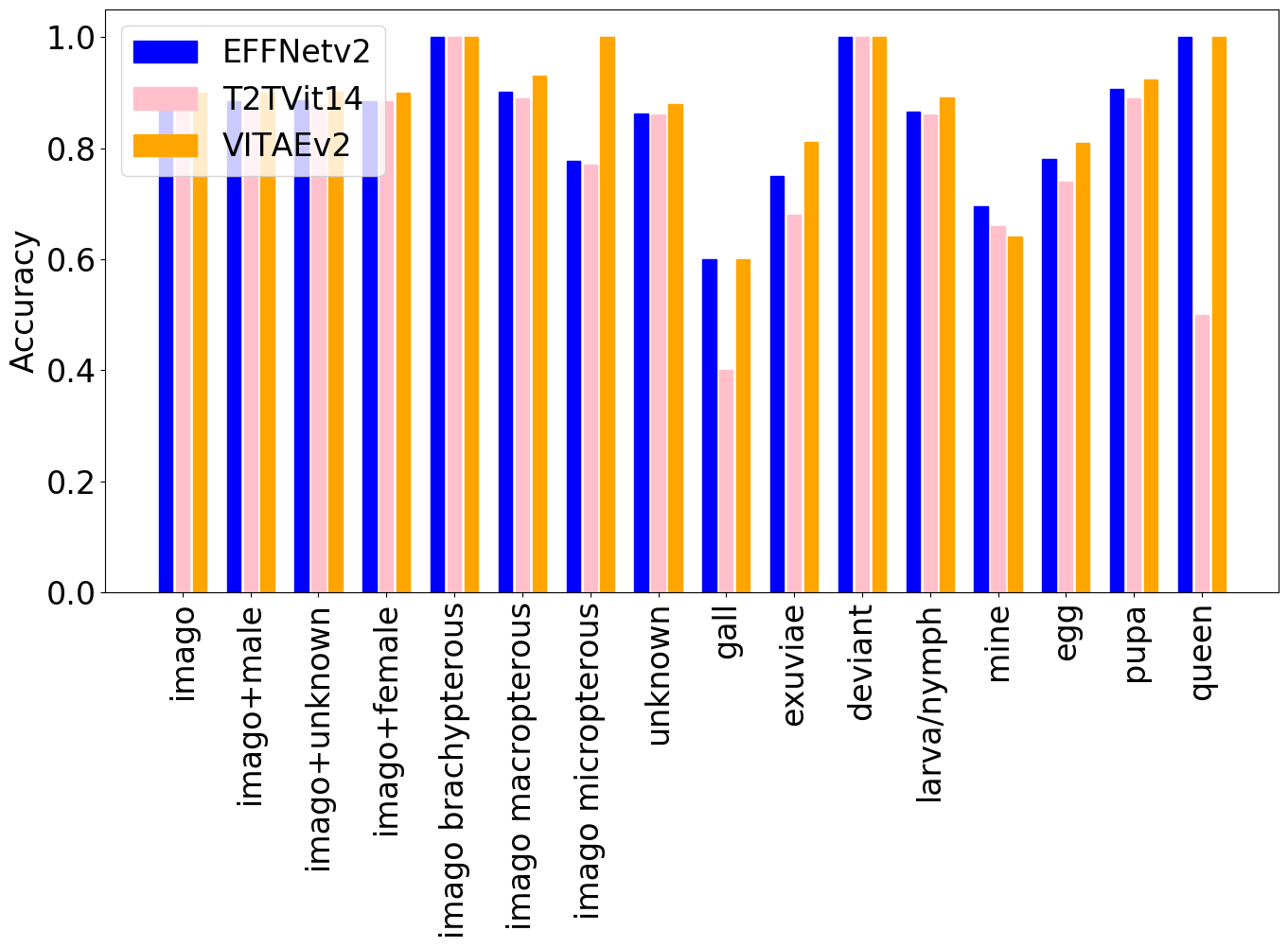}
  \caption{Coleoptera\_obs}
  \label{fig:Coleoptera_ms}
\end{subfigure}%
\begin{subfigure}{.5\textwidth}
  \centering
  \includegraphics[width=\linewidth]{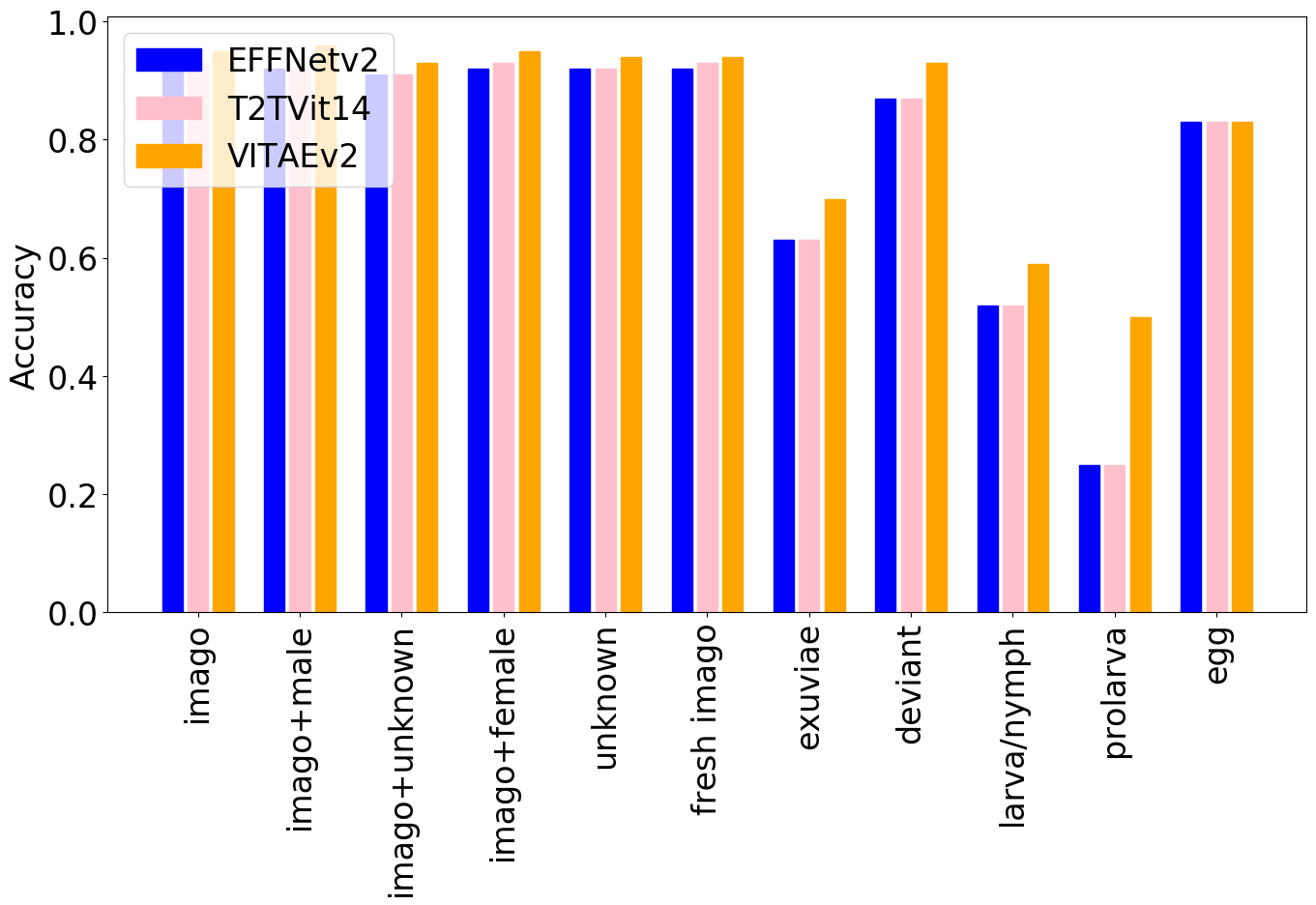}
  \caption{Odonata\_obs}
  \label{fig:Odonata_ms}
\end{subfigure}
\caption{The performance in top-1 accuracy in species prediction obtained for each morph.}
\label{fig:morphsex}
\end{figure*}
\begin{figure*}[!b]
\begin{subfigure}{.5\textwidth}
  \centering
  \includegraphics[width=\linewidth]{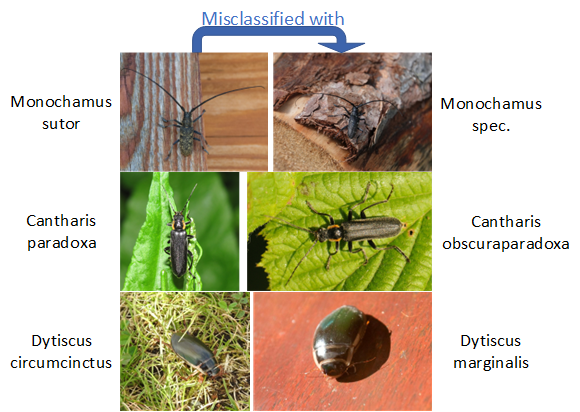}
  \caption{Coleoptera\_obs misclassified}
  \label{fig:Coleoptera_f}
\end{subfigure}%
\begin{subfigure}{.5\textwidth}
  \centering
  \includegraphics[width=\linewidth]{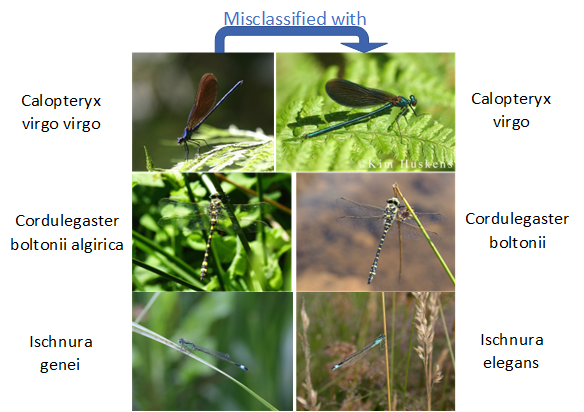}
  \caption{Odonata\_obs misclassified}
  \label{fig:Odonata_f}
\end{subfigure}
\caption{Some of the most common failures of all three models.}
\label{fig:failure}
\end{figure*}

The second level of fine-grained task considered is morph/sex, as discussed in Sec.~\ref{sec:datasets}, both Coleoptera and Odonata appear in nature in different morphs due to the different stages in life. This aspect affects the dataset by increasing the granularity of the type of images that the models need to learn intra-species. It is worth noting that the models are trained to identify the species, so the results as to be interpreted as the top-1 species prediction in relation to the morph and sex. The Coleoptera and Odonata have different numbers of morphs (Tab.~\ref{tab:samples_ms}), Fig.~\ref{fig:morphsex} shows the top-1 achieved by each model for each morph or sex in the case of imago morph. We observe that \myeffn and \myvitae both reach high performance with almost all the morphs in both the Coleoptera\_obs and the Odonata\_obs datasets. The \myvitae shows a similar behaviour as at the species level, in fact, it is able to well identify species in morphs that have a low number of samples. Moreover, for the imago morph, we evaluate the impact of sex on the performance of the models and we can conclude that all three models manifest a good performance with all three sexes.

\paragraph{\textbf{Understanding the failures}}
\label{par:failure}

In this section, we investigate the impact of misclassification to understand if they do make sense in biodiversity monitoring. We want to answer the question, are the models completely misunderstanding the species or there are some reasons that can justify the misclassification? To answer this question, we first consider the limitation of taxonomical datasets considered for this study. Datasets contain generic species refined to the actual species classification. This means that samples that are not identifiable by the taxonomist at the species level are labelled at the genus level with 'spec.' species. This peculiarity of the datasets generates some of the most common misclassifications in all three models. 
To prove this observation, we consider the species where more than $50\%$ of the test data are misclassified. We observe that all these species are misclassified within the superspecies, or a species from the same genus or the generic genus. Fig.~\ref{fig:failure} shows samples of these types of misclassification obtained with all three models. In Fig.~\ref{fig:Odonata_f}, both the first two species, \textit{Calopteryx virgo virgo} and \textit{Cordulegaster boltonii algirica}, are confused with their superclasses, \textit{Calopteryx virgo} and \textit{Cordulegaster boltonii} respectively. In both Fig.~\ref{fig:Coleoptera_f},~\ref{fig:Odonata_f}, \textit{Cantharis paradoxa}, \textit{Dytiscus circumsinstus}, and \textit{Ischnura genei} are confused with other species that belong to the same genus. Finally in Fig.~\ref{fig:Coleoptera_f}, the \textit{Monochamus sutor} is confused with the generic genus, the \textit{Monochamus spec.} (which is a class in the dataset that represents a generic species within the genus \textit{Monochamus}). These mistakes occur mostly with the species that are rare and so less represented in the datasets.
Therefore, these misclassifications make sense and open new discussions on the proper use of these methods in ecology to exploit the possibility of using such models to help taxonomist to identify difficult images at the species level. 
\section{Discussion}
\label{sec:discussion}
Our results demonstrate that the introduction of transformers provides fast models that outperform the CNN models in accuracy at the species level and the morph and sex level. For the fine-grained species level, all three models show an overall good performance. As shown in Fig.~\ref{fig:accSpecies}, \myvitae proves to be more robust and performs well on rare species, it has fewer species with $0\%$ top-1. We observe similar behaviour in the \myttv models even if with lower performance. The evaluation of the three models on iNaturalist and Artportalen sub-datasets, shows performances in line with the ones achieved with the Coleoptera\_obs  and the Odonata\_obs datasets. These results confirm the robustness of the models and their good generalization to different distributions. The \myeffn achieves competitive results that are slightly lower than the models based on transformers. 

It is worth noting the analysis of the fine-grained task at morph/sex level. The results show the ability of the models to extract generalised representations for the species considering the intra species dimorphism among the different stages in life. We need to point out that the models achieve high performance for almost every morph and sex (Fig.~\ref{fig:morphsex}) but we observe that for morphs such as gall and queen in Coleoptera, the \myvitae and \myeffn outperform the \myttv, and for prolarva in Odonata, only \myvitae shows a good performance in classification. These results together with the low number of $0\%$ top-1, discussed before, make us identify the \myvitae as a model able to deal with the shortage in data for the rare species and morphs. 

The inference speed gives a completely different evaluation of the models presenting the \myttv as the faster model compare to the other two models. Finally, we need to consider the end use of these models: if the performance and the robustness are the features mainly required, the \myvitae and \myeffn models are the most suitable for the fine-grained tasks with a preference for the \myvitae; if the focal point is for public use so the inference speed is to be considered, the \myttv demonstrated to achieve good performance faster than the others.

\section{Conclusions}
In this paper, we investigated three state of the art models for fine-grained tasks in the taxonomic domain. The focus was on the comparison among fully-convolutional, fully-transformer, and hybrid, to depict the strengths and the weakness of three families. Our analysis shows different trade-offs between performance and inference speed. Hybrid models can identify species well even in case a low number of samples is available for training but these models demand a long inference time. Fully-convolutional models obtain good performance but we observe a lower performance with rare species compared to the hybrid models. Finally, the fully-transformer models are slightly less performant than the hybrid ones, but these models are faster compared to the other two models considered. We conclude that all three choices remain as candidates for more corroborating studies and for future deployment.

\section*{Acknowledgement}
This research was supported by the EU Horizon projects MAMBO and TETTRIs.
\newpage

\bibliographystyle{plainnat}
\bibliography{manuscript.bib}
\end{document}